\documentclass[conference]{IEEEtran}
\IEEEoverridecommandlockouts
\usepackage{cite}
\usepackage{amsmath,amssymb,amsfonts}
\usepackage{algorithmic}
\usepackage{graphicx}
\usepackage{textcomp}
\usepackage{xcolor}
\usepackage{multirow}
\usepackage{adjustbox}
\usepackage[many]{tcolorbox} 
\usepackage{url}
\usepackage{subfig}
\usepackage{float}
\usepackage{calc}  
\usepackage{balance}
\usepackage{enumitem}  
\usepackage[]{mdframed}
\usepackage{hyperref}

\newtcolorbox{boxA}{
    fontupper = \bf,
    boxrule = 1.5pt,
    colframe = black % frame color
}

\def\BibTeX{{\rm B\kern-.05em{\sc i\kern-.025em b}\kern-.08em
    T\kern-.1667em\lower.7ex\hbox{E}\kern-.125emX}}
\begin{document}

\title{Tapping in a Remote Vehicle's onboard LLM to Complement the Ego Vehicle's Field-of-View\\

}

\author{Malsha Ashani Mahawatta Dona$^1$, Beatriz Cabrero-Daniel$^1$, Yinan Yu$^2$, Christian Berger$^1$\\
$^1$\textit{University of Gothenburg} and $^2$\textit{Chalmers University of Technology}\\
%Department of Computer Science and Engineering\\
Gothenburg, Sweden \\
\{malsha.mahawatta,beatriz.cabrero-daniel,christian.berger\}@gu.se, yinan@chalmers.se}

\maketitle

\begin{abstract}

Today's advanced automotive systems are turning into intelligent Cyber-Physical Systems (CPS), bringing computational intelligence to their cyber-physical context. Such systems power advanced driver assistance systems (ADAS) that observe a vehicle's surroundings for their functionality. However, such ADAS have clear limitations in scenarios when the direct line-of-sight to surrounding objects is occluded, like in urban areas. Imagine now automated driving (AD) systems that ideally could benefit from other vehicles' field-of-view in such occluded situations to increase traffic safety if, for example, locations about pedestrians can be shared across vehicles. Current literature suggests vehicle-to-infrastructure (V2I) via roadside units (RSUs) or vehicle-to-vehicle (V2V) communication to address such issues that stream sensor or object data between vehicles. When considering the ongoing revolution in vehicle system architectures towards powerful, centralized processing units with hardware accelerators, foreseeing the onboard presence of large language models (LLMs) to improve the passengers' comfort when using voice assistants becomes a reality. We are suggesting and evaluating a concept to complement the ego vehicle's field-of-view (FOV) with another vehicle's FOV by tapping into their onboard LLM to let the machines have a dialogue about what the other vehicle ``sees''. Our results show that very recent versions of LLMs, such as GPT-4V and GPT-4o, understand a traffic situation to an impressive level of detail, and hence, they can be used even to spot traffic participants. However, better prompts are needed to improve the detection quality and future work is needed towards a standardised message interchange format between vehicles.

\end{abstract}


\section{Introduction}

% The new introduction starts here - MM
% Concept of Cyber-Physical Systems (CPS) in automotive environments.
In recent years, advanced technologies have been integrated into vehicles turning them into intelligent Cyber-Physical Systems (CPS) to improve comfort and safety. The development of such intelligent systems aims to enhance safety and efficiency while providing an overall improved driving experience.  
CPS represents a paradigm where computational platforms are integrated with control algorithms, which are capable of handling heterogeneous distributed systems \cite{chakraborty2016automotive} as we face them in automotive systems. CPS uses computational resources to interact with physical processes; modern vehicles, for example, perceive their surroundings with cameras, radars and ultrasound devices and even road-side units (RSUs) to act safely in their context.

% Role of LLMs as a communication interface for CPS within vehicles.
Current technological advancements potentially influencing automotive systems and CPS include Large Language Models (LLMs) that leverage natural language processing (NLP) techniques to understand and respond even to complex multi-modal data inputs. LLMs such as Generative Pre-trained Transformers (GPT) excel in various domains due to their exceptional language understanding and generation capabilities~\cite{generativeAI_progress}. Hence, they have shown potential applicability in various domains such as health care, education, research, etc.~\cite{healthcare11060887} to provide better experiences and additional services to the users efficiently.

% However, common to the aforementioned ADAS is not only a clearly defined ODD but also that other traffic participants, with which these systems interact like other vehicles or pedestrians, are directly in the line-of-sight of the own vehicle.

\subsection{Problem Domain and Motivation}
%CBe: OK
For today's ADAS to interact with other traffic participants, such as other vehicles or pedestrians within the defined operational design domains (ODDs), they typically need to be within a sensor's field of view (FOV). When relaxing the constraint of requiring other traffic participants to be in a vehicle's field-of-view (FOV) to explore opportunities and challenges beyond today's generation of ADAS or even towards automated driving (AD), technical concepts typically suggest either the use of V2I via RSUs or to share sensor data or object information between vehicles by streaming data from one vehicle to another one (V2V) to complement a vehicle's own field-of-view~\cite{RSU_onboardDataTransfer}. However, RSUs do not see a broad availability to support such scenarios, and streaming other vehicles' data is consuming a substantial amount of cellular network bandwidth while only a fraction of the information in the streamed data is of relevance for an ADAS or AD system.

% CBe: OK
With the recent advancements in the automotive context, as, for example, seen in the SOAFEE framework \cite{soaffee}, vehicle system architectures are trending towards powerful centralized processing units that include CPUs and GPUs. Anticipating the presence of an LLM in such a vehicle could be used to improve, for instance, the passengers' in-car experience with better voice assistants processing even complex natural language-based dialogues between the human and the vehicle. In fact, some global automotive manufacturers are already working on deploying state-of-the-art foundational models within the vehicles to assist the in-car experience for passengers \cite{bmwCarExpert,bmwWebsite}.

\begin{figure}
    \centering
    \includegraphics[width=\linewidth,scale=0.25]{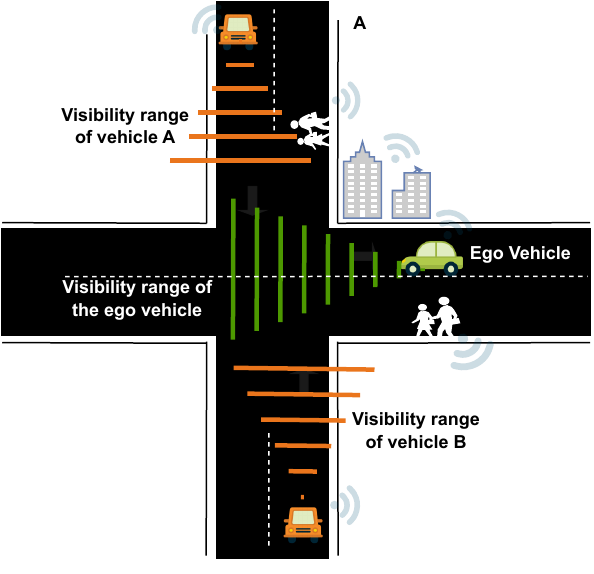}
    \caption{Graphical representation of a looking around the corner problem: The ego vehicle is approaching an intersection together with vehicles A and B. Each vehicle has its own FOV represented by their respective colours, and that information could help each other understand the complete traffic situation at hand.}
    \label{fig:intersection}
\end{figure}

% CBe: OK
We suggest a concept to exploit the anticipated availability of such technologically advanced smart vehicles to circumvent the aforementioned relaxed constraints to enrich the own vehicle's FOV for practical traffic situations such as looking around the corner at intersections~\cite{lookingaroundthecorner}: As shown in Fig.~\ref{fig:intersection}, the ego vehicle is approaching an intersection where other vehicles A and B are in proximity as well. All vehicles have not entered their respective line-of-sight and hence, information about their own surroundings could help the other vehicles to better understand the traffic situation. Instead of letting vehicles A and B stream their data to the ego vehicle, or being dependent on an RSU for distributing sensor data between vehicles, we envision that the ego vehicle is having a \emph{dialogue} with other vehicles via their respective LLMs to enrich its own understanding of the surrounding traffic.
% CBe: OK

Compared to streaming images over a VANET with 1Mbps network link including 10\% overhead, a 218.6KB image would require 1.98 seconds to complete the transmission. In contrast, the same amount of data corresponds to a dialog fitting on approx.~124 pages of text. Hence, we can assume that even a detailed back-and-forth dialogue would require less data compared to transmitting images. Therefore, we envision that our proposed approach will bring tremendous advantages over the current technical concepts suggested in the scientific literature: (A) the amount of data to be shared can be reduced substantially as only small texts are exchanged as we will illustrate and evaluate in our study; (B) no intellectual property (IP)-sensitive information about hardware capabilities of the data-supplying vehicles need to be exchanged as the LLM is acting as a communication interface within the CPS abstracting such details; and (C) no IP-sensitive information about the software or machine learning (ML) models such as ML-weights need to be exchanged.

\subsection{Research Goal and Research Questions}
\label{sec:researchQuestions}
Our main research goal is to evaluate the effectiveness of using an LLM to serve as a communication interface between vehicles to detect pedestrians that are not in direct line of sight. We address the following research questions: 

\begin{description}[leftmargin=!,labelwidth=\widthof{\bfseries RQ-1:}]
\item[RQ-1:] To what extent can a state-of-the-art LLM be used as a communication interface to qualitatively identify pedestrians?
\item[RQ-2:] To what extent can a state-of-the-art LLM be used to reliably detect the location of pedestrians?
\end{description}

\subsection{Contributions and Scope}

% CBe: OK
We explore the possibility of applying an LLM for pedestrian detection by conducting a set of experiments. Our main contribution is a qualitative assessment of a state-of-the-art LLM to detect pedestrians while using a minimalistic dialogue for a curated dataset containing single pedestrians on a crosswalk. While our experiments unveil impressive fine-granular details, the LLM's quality to locate a pedestrian as well as the execution times need to be improved further.

\subsection{Structure of the Paper}

% CBe: OK
The remainder of the paper is organised as follows: In Section~\ref{sec:relatedWork}, we review the related work. Section~\ref{sec:methodology} provides a detailed description about the adopted methodology that addresses the research goal and the research questions for our study. In Section~\ref{sec:results}, we present our results in detail, and Section \ref{sec:discussion} provides the analysis and discussion of the findings. We conclude the paper in Section\ref{sec:conclusion}. 

\section{Related Work}
\label{sec:relatedWork}

When we look at the feasibility of deploying LLMs within a CPS for better utilization of resources in handling complex tasks, studies such as Xu et al.~\cite{xu2024penetrative} and Yang et al.~\cite{yang2024llmbased} show the potential of using LLMs to handle various tasks in dynamic environments. Xu et al.~coined a concept called ``Penetrative AI'', introducing LLMs to interact with and reason about the physical environment through various types of sensors \cite{xu2024penetrative}. The study's findings showcase that LLMS such as ChatGPT are proficient in interpreting IoT sensor data and reasoning about them according to the tasks in the physical world. Furthermore, the study of Yang et al.~also demonstrate that LLMs can be employed in physical environments as a dynamic solution, illustrating the potential of integrating LLMs into CPS to enhance their productivity \cite{yang2024llmbased}. These studies motivate delving deeper into the concept of integrating LLMs with CPS. 

The problem of enriching a vehicle's FOV as in scenarios like looking around the corner~\cite{lookingaroundthecorner} has been addressed for a while to find an effective and feasible solution that can be applied in Cooperative Intelligent Transportation Systems (C-ITS). Isele et al.~\cite{RelatedWork_1_DavidIsele} explored the effectiveness of Deep Reinforcement Learning (RL)-based approaches for intersection handling. They report that Deep Q-Network-based approaches show better task efficiency and success rates compared to traditional rule-based methods. The authors collect a point cloud through a combination of six Light Detection and Ranging (LiDAR) sensors from an autonomous vehicle to simulate a real-world unsignaled T-junction scenario in SUMO to test the Deep-Q-Network-based approaches and traditional Time-To-Collision (TTC) algorithms. However, in the event of occlusions within the intersection, the deep Q-networks resulted in collisions. 

Kilani et at.~\cite{RelatedWork_2_Kilani} propose a framework to improve the driver's FOV to detect road obstacles using LiDAR point cloud data. The proposed method can be used to identify obstacles such as buildings, vegetation, or any other roadside infrastructure that occludes the driver's FOV. The latter part of the proposed method investigates how visibility assessment data can be used in achieving intersection safety by mitigating collisions. The study shows that the higher blockage rates increase the collision proportions indicating a higher risk of collisions in intersections.

Zhou et al.~\cite{RelatedWork_3_Zhou} present a framework called EAVVE as a vehicle-to-everything system that utilises edge servers to provide vehicular vision to mitigate collisions. The authors have evaluated the framework against real-world road testing in different traffic densities using different infrastructures. The prototype of EAVVE works on positional data and images captured by sender vehicles in a given geographical area. The image data will be processed through object detectors that are located in either the sender vehicle itself or at the edge of the network, like in RSUs. The detection results and the direct positional data will be used to provide real-time emergency detection and notifications to the client vehicles. EAVVE has shown significant performances in both real-world testing scenarios as well as in simulation-based environments.  

The previously mentioned studies have contributed to addressing the challenge of occluded traffic situations. However, respective limitations and gaps still remain to cover various traffic scenarios in the context of heterogeneous system architectures for vehicles. 

\section{Methodology}
\label{sec:methodology}

Our methodology to address the overall research goal is comprised of several parts that we describe in detail in the following. We included a preliminary exploratory study to determine the general feasibility of our idea to identify (RQ-1) and to locate (RQ-2) pedestrians with today's state-of-the-art LLMs, as technical advancements happen regularly and rapidly in this area. Afterward, we decided on a dataset that we used for the subsequent systematic experiments. These experiments entailed to determine to what degree LLMs can identify pedestrians at all and sub-experiments to determine the quality of an LLM's response to locate a pedestrian reliably.

\subsection{Preliminary Exploratory Studies}
\label{sec:preStudies}

\begin{figure}
    \centering
    \includegraphics[width=\linewidth]{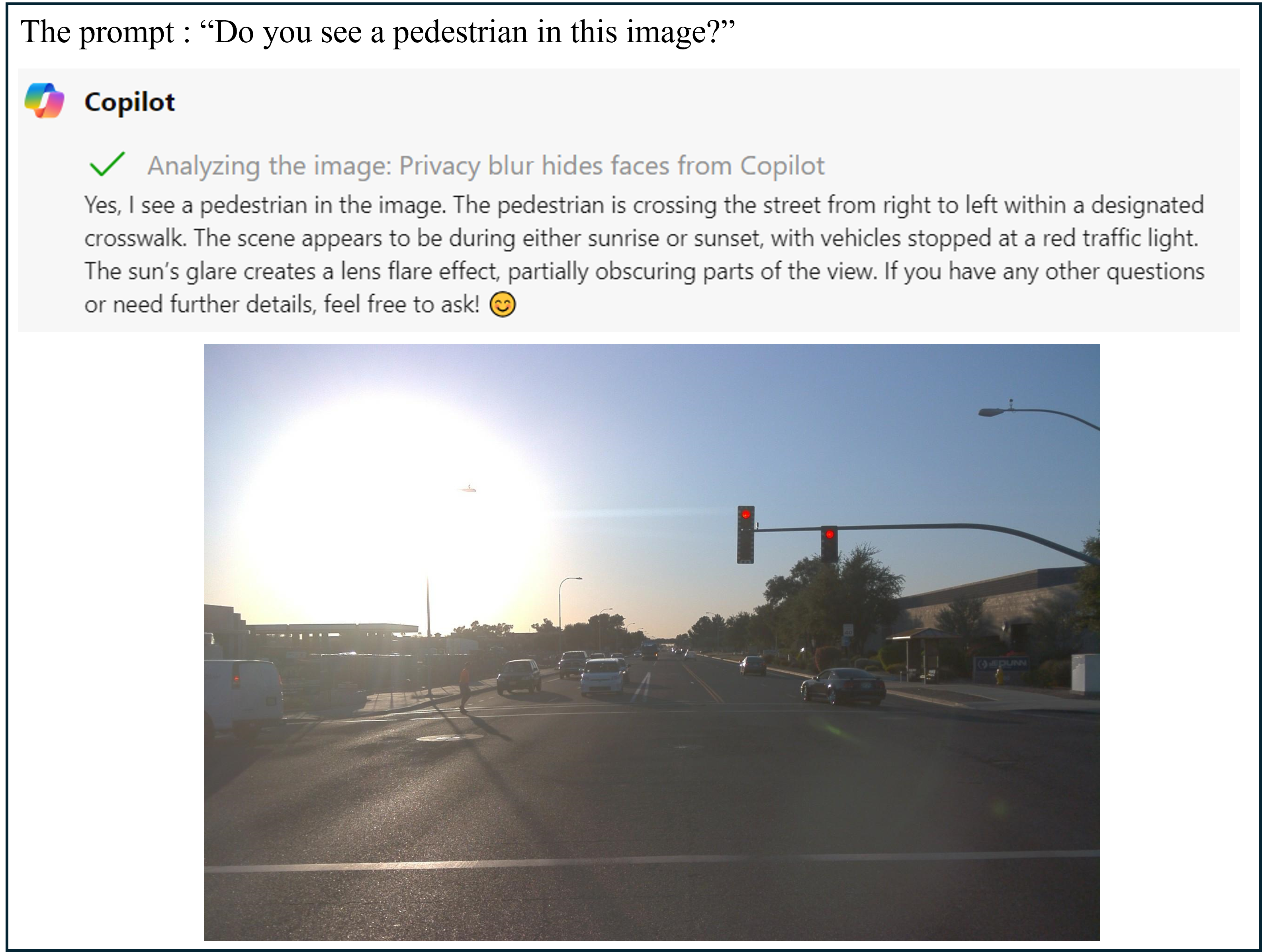}
    \caption{Detailed answers obtained from Microsoft Copilot. The question ``Do you see a pedestrian in this image?'' was used as the prompt. The input image is taken from the Waymo dataset~\cite{waymoDataset} and represents a sunset.}
    \label{fig:copilotresults1}
\end{figure}

We conducted a series of preliminary feasibility studies with Microsoft Copilot, DALL.E 2 and GPT-4. Our initial studies unveiled that using generative AI models for object detection is working impressively well already in a zero-shot setting \cite{zeroshot}. Fig.~\ref{fig:copilotresults1} shows a detailed description generated by Microsoft Copilot as an answer for the prompt ``Do you see a pedestrian in this image?''
However, localising the pedestrian posed a challenge for Microsoft Copilot, resulting in unwanted hallucinations \cite{hallucinations} occurring unpredictably from time to time. 

\begin{figure}
    \centering
    \includegraphics[width=\linewidth]{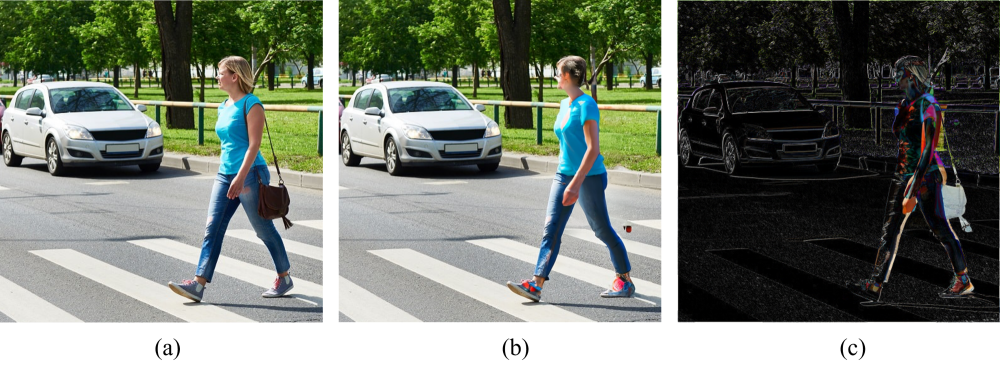}
    \caption{Exploring an LLM's capabilities not only to describe but also to locate a pedestrian: (a) shows the input image (taken from~\cite{womancrossingroad_reference}), (b) the DALL.E 2 generated image (b), and differences between both.}
    \label{fig:dall2_result1}
\end{figure}

We investigated as part of our pre-studies DALL.E 2 with the goal of to what extent DALL.E 2 is capable of not only detecting but also locating a pedestrian by generating a properly annotated image as a response. The difference between the original image supplied to the LLM and its generated response was calculated and analysed to determine to what degree the LLM could locate a pedestrian. We observed that the image edit feature of DALL.E 2 works best when supplying an image mask to guide the LLM where to search for the pedestrian as shown in Fig.~\ref{fig:dall2_result1}. While experimenting with different image masks seems to enhance an LLM's response, this approach was not further pursued as it is contrary to our research goal to let the LLM locate a pedestrian on its own where the requesting vehicle has no information at all where the pedestrians and other vehicles would be. We completed our preliminary exploratory studies with the conclusion that only GPT-4 with vision (GPT-4V) introduced in September 2023 \cite{openai2023gpt4v_paper} and GPT-4o (``o'' for ``omni'') introduced in May 2024 \cite{gpt4oWebsite} are able to detect objects with promising results.

\subsection{Datasets}
\label{sec:datasets} 

For our main experiments, we used the Waymo open dataset \cite{waymoDataset} and, in particular, the labelled objects on pedestrians. The Waymo dataset has been collected in urban and suburban areas in the USA in 2021, covering over 100,000 different day and night scenarios. The dataset was extracted with a sample rate of 10 Hz, focusing on the front camera images in the training set. The extraction contained 15,947 images, out of which the first 5000 images were considered to curate our dataset. 

The curated dataset contains 126 images with pedestrian labels to control potentially influencing factors for our experiment. We selected images having only one single pedestrian on a crosswalk in the centre part of the image. Another subset of 138 images was selected where pedestrians were not visible on crosswalks as a control group. These two sets of images were used throughout all our main experiments as described below. We used the VGG Image Annotator \cite{dutta2019vgg} to retrieve the bounding box coordinates of pedestrians that we later used as the ground truth for evaluation.

\subsection{Experiment Design}
\label{sec:experimentDesign}

After assessing the preliminary results, we decided to choose the recent release of GPT-4 with vision, also known as GPT-4V or gpt-4-vision-preview, and GPT-4o for all three experiments as these two models could analyse and reason about images in a promising manner. The obtained results will allow a comparison between two state-of-the-art LLMs. 

\subsubsection{Experiment 1: Binary Pedestrian Detection} \label{sec:methodexp1} \hfill \break

We designed this experiment to address RQ-1 by assessing to what extent the LLM can detect pedestrians. We used our curated dataset with the GPT-4V and GPT-4o models and the following prompt:

\begin{boxA}
Is there a human pedestrian in this image? Answer only either ``yes'' or ``no''. 
\end{boxA}

\subsubsection{Experiment 2: Bounding Box Generation} \hfill \break
\label{sec:experiment2}

We designed this experiment to address RQ-2 assessing to what extent GPT-4V and GPT-4o can locate a pedestrian using a prompt that requests the LLM to return the coordinates of the area occupied by the pedestrian. Details about the reference coordinate system to be used were fed to the models along with the prompt to standardise the coordinate system of pedestrian detection. All images in the curated dataset were checked against the prompt. These coordinates generated by GPT models were compared together with the ground truth coordinates retrieved from the Waymo Dataset. \newline

\begin{boxA}
Given the reference system where (0,0) is the top-left corner and (1,1) is the bottom-right corner of the image, provide the coordinates (X,Y), (X',Y') representing the precise location of a person in the image. Ensure the coordinates accurately delineate the complete area occupied by the person. Return coordinates ONLY using this template: (X,Y), (X',Y')
\end{boxA}

\subsubsection{Experiment 3: Comparative Evaluation of Prompts} \hfill \break
\label{sec:experiment3}

We expanded the previous experiment by systematically checking the curated dataset three times against the three prompts listed in Tab.~\ref{tab:prompts} to conduct the comparative evaluation across prompts. The multiple runs across the same prompt facilitated analysing the consistency of the responses to identify potential hallucinations as suggested in literature~\cite{ronanki2022chatgpt}. The initial prompt used in experiment 2 was refined repeatedly based on the feedback of a generative AI model. The generated bounding box details, respective ground truth coordinates, and processing times were recorded for each prompt. Furthermore, the results recorded under both models were compared against each other for performance evaluation. 

\renewcommand{\arraystretch}{1.4} % Default value: 1

\begin{table*} [ht]
    \centering
    \caption{List of prompts that were used in the Experiment 3} \label{tab:prompts}
    \resizebox{\linewidth}{!}{
    \begin{tabular}{|c|p{17cm}|} \hline  
        \textbf{P1} & Given the reference system where (0,0) is the top-left corner and (1,1) is the bottom-right corner of the image, provide the coordinates (X,Y), (X',Y') representing the precise location of a person in the image. Ensure the coordinates accurately delineate the complete area occupied by the person. Return coordinates ONLY using this template: (X,Y), (X',Y') \\ 
        \hline  
        \textbf{P2} & In the image, identify the person's location using a coordinate system where (0,0) is the top-left corner and (1,1) is the bottom-right corner. Provide the coordinates (X,Y), (X',Y') that encapsulate the entire area occupied by the person. Use this format only: (X,Y), (X',Y') \\ \hline  
        \textbf{P3} & Using (0,0) is the top-left corner and (1,1) is the bottom-right corner, provide the coordinates (X,Y), (X',Y') of the location of a person in the image. The coordinates should contain the complete area occupied by the person. Return coordinates ONLY using the template: (X,Y), (X',Y') \\ \hline   
    \end{tabular}     
    }
 \end{table*}

\section{Results} \label{sec:results}

We report the results from the three experiments in the following. For replication purposes, we share the necessary code and raw results as Supplementary Material in a GitHub repository~\cite{LLMrepogithub}.

\subsection{Experiment 1: Binary Pedestrian Detection} \label{sec:resultsexp1}

% WE WANTED TO BINARY LABEL PEDESTRIAN APPARITIONS
In this experiment, GPT-4V and GPT-4o both were prompted using the images from our curated subset. The two models were then asked to determine whether a pedestrian was present in each of the individual images. The generations were then contrasted to the manual labels for each of the images, and the instances where the models correctly predicted the labels were counted.

% GPT WORKS VERY WELL
As shown in Tab.~\ref{tab:exp1confusion}, both models GPT-4V and GPT-4o can correctly identify the pedestrians in all images (high recall) at the expense of wrongly labelling images with no pedestrians (false positives). The trade-off between recall and other metrics, as reported in Tab.~\ref{tab:exp1confusion}, aligns with the goal of enriching a vehicle's FOV to reliably detect pedestrians when they are indeed present. 

% DIFFERENT RUNS
The same confusion matrix was obtained when re-running the experiment. However, it is important to note that even though none of the values for the metrics in Tab.~\ref{tab:exp1confusion} changed, the false positives were occurring in different images in the different runs.

The Tab.~\ref{tab:statistics} contains a comparative evaluation between the two models GPT-4V and GPT-4o to report about recall, specificity, and precision.  

\begin{table*}
    \centering
    \caption{Confusion matrix and relevant performance metrics for GPT-4V and GPT-4o in pedestrian detection over the dataset.} \label{tab:exp1confusion}
\begin{tabular}{c|cc|c}
\cline{2-3}
% \multirow{2}{*}{\textbf{}} & \multicolumn{2}{c|}{\textbf{Ground Truth}} &  \\ \cline{2-3}
 & \multicolumn{1}{c|}{\textbf{Ground Truth: True}} & \textbf{Ground Truth: False} &  \\ \hline
\multicolumn{1}{|c|}{\textbf{GPT-4V Predicted True}} & \multicolumn{1}{c|}{120 (TP)} & 5 & \multicolumn{1}{c|}{Recall$_{4V}$ = 95.24\%} \\ \hline
\multicolumn{1}{|c|}{\textbf{GPT-4V Predicted False}} & \multicolumn{1}{c|}{6} & 129 (TN) & \multicolumn{1}{c|}{Specificity$_{4V}$ = 96.27\%} \\ \hline 
\multicolumn{1}{|c|}{\textbf{GPT-4o Predicted True}} & \multicolumn{1}{c|}{125 (TP)} & 11 & \multicolumn{1}{c|}{Recall$_{4o}$ = 99.21\%} \\ \hline
\multicolumn{1}{|c|}{\textbf{GPT-4o Predicted False}} & \multicolumn{1}{c|}{1} & 127 (TN) & \multicolumn{1}{c|}{Specificity$_{4o}$ = 92.03\%} \\ \hline
 % & \multicolumn{1}{c|}{Precision = 96.00\%} & Negative Predictive Value = 95.56\% & \multicolumn{1}{c|}{Accuracy = 95.77\%} \\ \cline{2-4}
\end{tabular}
\end{table*}

% TP 120
% FN 6
% FP 5
% TN 129
% Sensitivity 0.9524
% Specificity 0.9627
% Precision 0.9600
% Negative Predictive Value 0.9556
% False Positive Rate 0.0373
% False Discovery Rate 0.0400
% False Negative Rate 0.0476
% Accuracy 0.9577
% F1 Score 0.9562
% Matthews Correlation Coefficient 0.9153

\begin{table}[]
    \centering
    \begin{tabular}{c|cc}
    \textbf{Statistic} & \textbf{GPT-4V} & \textbf{GPT-4o} \\ \hline
Recall & 95.24\% & \textbf{99.21\%} \\
Specificity & \textbf{96.27\%} & 92.03\% \\
Precision & \textbf{96.00\%} & 91.91\% \\
Negative Predictive Value & 95.56\% & \textbf{99.22\%} \\
False Positive Rate & 3.73\% & \textbf{7.97\%} \\
False Discovery Rate & 4.00\% & \textbf{8.09\%} \\
False Negative Rate & \textbf{4.76\%} & 0.79\% \\
Accuracy & \textbf{95.77\%} & 95.45\% \\
F1 Score & \textbf{95.62\%} & 95.42\% \\
Matthews Correlation Coefficient & \textbf{91.53\%} & 91.18\% \\
    \end{tabular}
    \caption{Comparative evaluation of GPT-4V and GPT-4o}
    \label{tab:statistics}
\end{table}
 
% \begin{table}
%     \centering
%     \caption{Metrics for the performance of the GPT-based pedestrian detector.} \label{tab:exp1metrics}
%     \begin{tabular}{|c|c|} \hline  
%         \textbf{Metric} & \textbf{Value} \\ \hline  
%         Recall & 100\% \\ \hline   
%         Specificity & 72.29\% \\ \hline   
%         Precision & 70.51\% \\ \hline  
%         Negative Predictive Value & 100\% \\ \hline  
%         F1-score & 82.71\% \\ \hline 
%     \end{tabular}     
%  \end{table}

% GPT DOES KNOW WHERE PEOPLE ARE
Even though the OpenAI API documentation~\cite{gpt4v} states that GPT-4V is not ready to accurately answer questions about the positions of objects, our preliminary tests showed that it can understand whether a pedestrian is present, including detailed natural language explanations of where the pedestrian is and where they are headed to. This natural language explanation capability was also shown by GPT-4o, the most recent release of GPT-4 model.  For instance, about the image used in Fig.~\ref{fig:SummaryOfBahaviours_Permodel_perRun_perprompt}, GPT-4V stated that ``The individual is located in the centre of the image, standing in the middle of a crosswalk on the road [\dots] The position of the pedestrian is directly between the two lanes of the road'' whereas GPT-4o stated that ``The pedestrian is in the middle of a crosswalk, facing to the right, and holding a ``STOP'' sign.'' Similarly, detailed descriptions of the position and predictions of the intentions of the pedestrians were obtained from other images support the use of LLMs to obtain precise information about traffic.

\subsection{Experiment 2: Bounding Box Generation} \label{sec:resultsexp2}

% \textcolor{brown}{This is a draft. Please, focus on the contents and the succession of ideas, not the form. - Bea}

\begin{figure}
    \centering
    \includegraphics[width=\linewidth]{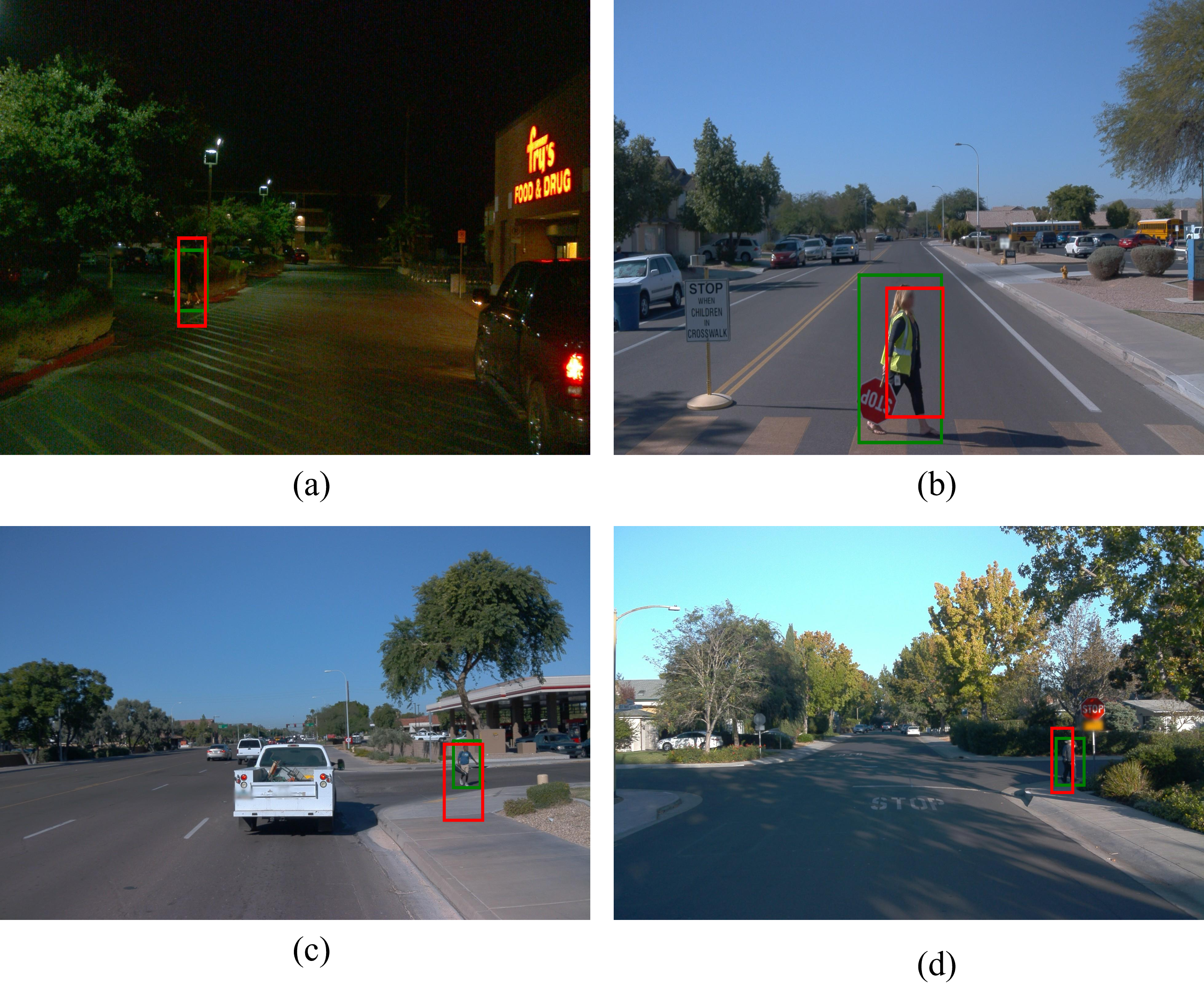}
    \caption{Images from Waymo Dataset\cite{DatasetReference} with a GPT-4V and GPT-4o generated bounding box (red) almost completely covering the ground truth area (green). The selected images show day and night scenarios. The results (a) and (b) are retrieved from the GPT-4V model, and the (c) and (d) are retrieved from the GPT-4o model.}
    \label{fig:highrecallImageSetResult}
\end{figure}

% SO WE ASKED IT FOR COORDINATES AND STUDIED THE PERFORMANCE
The second experiment further explored the ability of the models to locate the identified pedestrians, addressing \textbf{RQ-2}, by asking the LLM to provide coordinates using a specific reference system. To understand the overall performance, the following measures based on the differences between the generated and the ground truth bounding boxes were used: 
\begin{itemize}
    \item Number of unions: Percentage of images, in which the generated bounding box overlaps with the manual bounding box (as seen in Fig.~\ref{fig:highrecallImageSetResult}).
    \item Recall: Percentage of the manual bounding box that overlaps with the generated one (Fig.~\ref{fig:highrecallImageSetResult} shows an example with recall close to 100\% while the images that do not share any overlapping between the manual bounding box and generated one reported 0\% recall).
    \item Intersection over union: Relation between the overlapping between the generated and manual bounding boxes and the intersection between the two. This metric penalises big bounding boxes that would achieve 100\% recall.
\end{itemize}

% HOWEVER, GPT DOES IT POORLY
We computed these metrics for all images across three runs, and we obtained that the generated bounding boxes overlapped with the manually annotated ones in only 30.42\% of the tests with GPT-4V, whereas it was recorded as 42.18\% for GPT-4o. In the cases where the manual and generated bounding boxes did overlap, the recall was recorded as 37.38\% and 49.76\% for both GPT4-V and GPT-4o. This means that only around a third of the area occupied by a pedestrian is covered, on average, in only a third of the images containing a pedestrian.

\begin{figure}
    \centering
    \includegraphics[trim={0 0 0 2.5cm},clip, width=\linewidth]{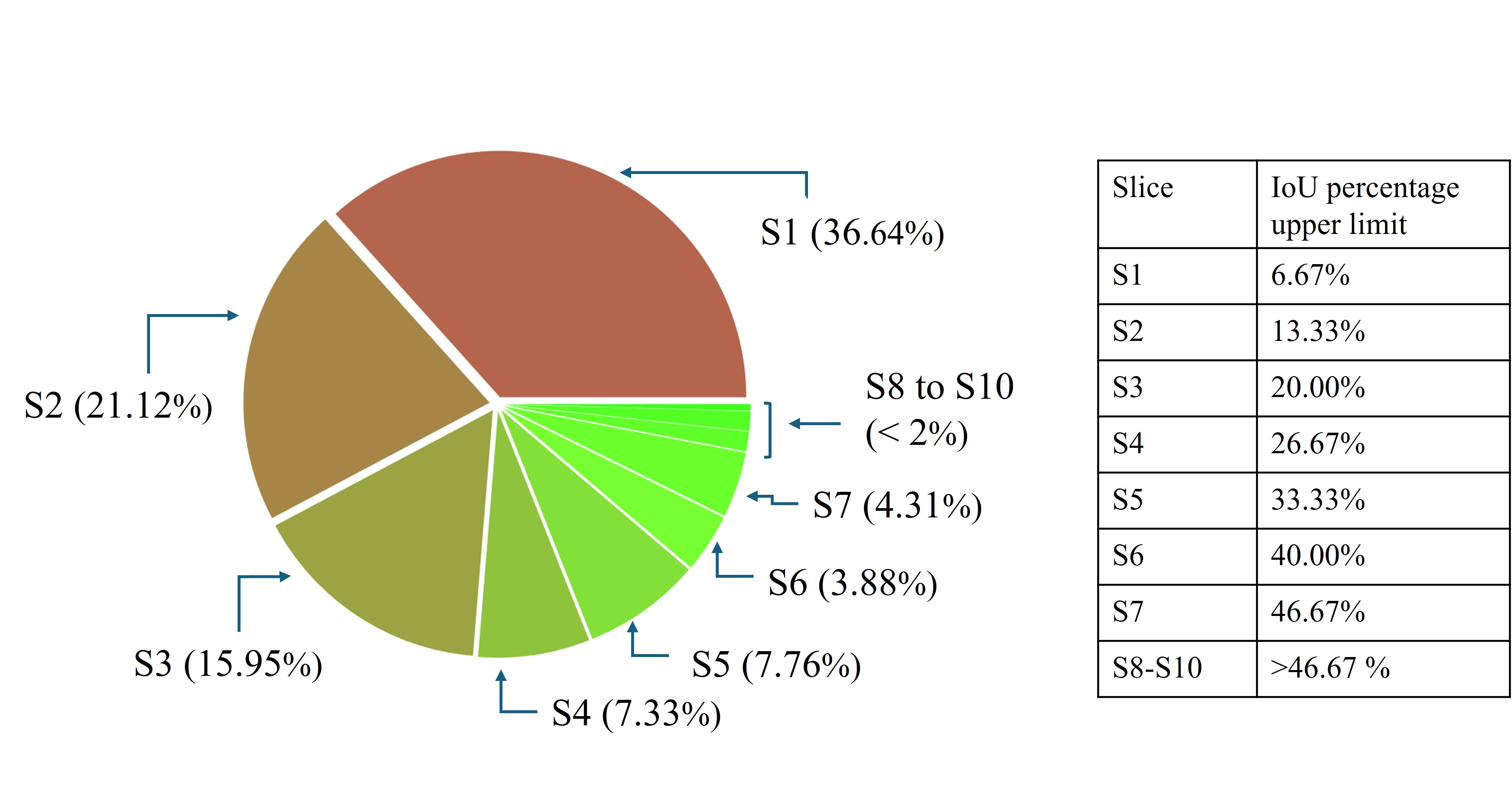}
    \caption{Intersection-over-Union (IoU) percentages for the images that share an overlapping between the GPT-4V generated and the ground truth bounding boxes, zooming in the overlapping IoUs. The pie chart contains 15 slices.}
    \label{fig:intersectionoverunion}
\end{figure}

% AND IT VARIES A LOT ACROSS IMAGES, WHICH MEANS IT IS INTERESTING TO STUDY WHY (FUTURE WORK)
Moreover, the standard deviation across images and runs was 28.81\% and 35.77\% for both models, respectively, which is substantial. This means that there may be some images where the overlap is bigger, as in Fig.~\ref{fig:highrecallImageSetResult}, and some images where there is very little to no overlap. To explore this variance, we analysed the performance of GPT-4V across images. The distribution of the recall for all images in the dataset containing a pedestrian can be seen in the left-most box in Fig.~\ref{fig:promptcomparison}. Fig.~\ref{fig:intersectionoverunion} represents the percentages of Intersection over Union (IoU) of the bounding boxes generated by GPT-4V considering the images where an overlapping was reported. However, the average IoU in all images is smaller. This is due to taking into account the area of the generated bounding box and penalises too large areas (covering the whole picture would lead to a 
recall of 100\%, but it would not be informative).

\begin{figure}
    \centering
    \includegraphics[width=\linewidth]{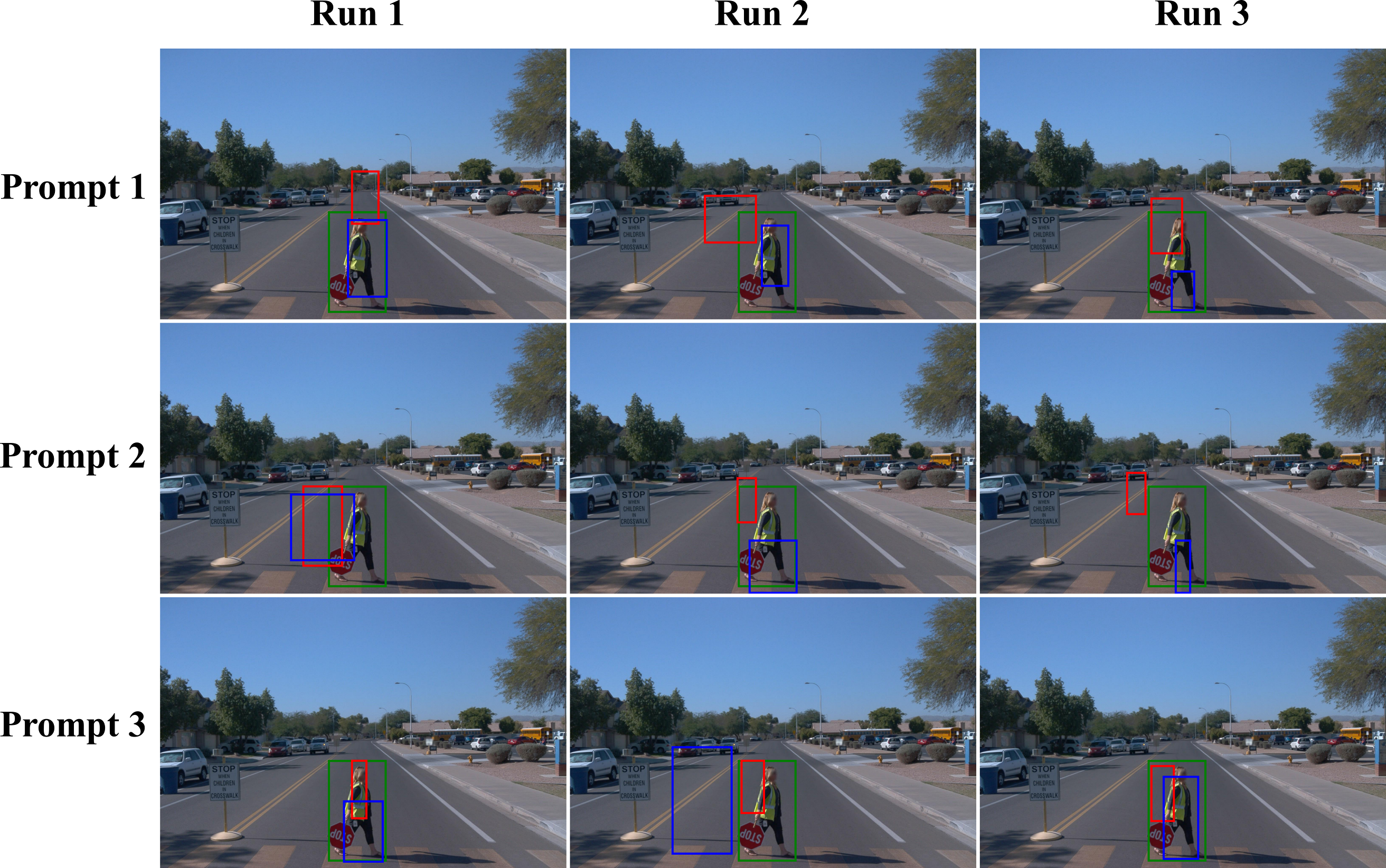}
    \caption{The behaviour of models GPT-4V (blue), GPT-4o (red) against the ground truth (green) in multiple runs across each prompt}
    \label{fig:SummaryOfBahaviours_Permodel_perRun_perprompt}
\end{figure}

% IT IS INTERESTING TO STUDY WHY (FUTURE WORK)
The individual test results, as seen in Fig.~\ref{fig:SummaryOfBahaviours_Permodel_perRun_perprompt} show that the performances of both models are not consistent across runs, which makes the LLMs unreliable (low recall) out of the box, as we further discuss in Sec.~\ref{sec:discussion}. The first row in Fig.~\ref{fig:SummaryOfBahaviours_Permodel_perRun_perprompt} depicts the behaviour of both models for the prompt considered in Experiment 2. Further analysis of the results also shows that some of the images are indeed challenging for the LLM. We conducted subsequent data analysis to figure out the reasons behind the incorrect responses produced by both the LLMs. The three main categories of error messages were identified as (A) No pedestrians detected, (B) Partial coordinates and (C) Ambiguous descriptions.

The majority of the error messages caused by GPT-4V fell under the partial coordinates category, where the constraint of the maximum token list may have hindered the generation of the response. However, had the model been behaving according to the prompt and only providing the coordinates, the selected maximum token count would have been sufficient. GPT-4o provided more ``no pedestrian detected'' error messages compared to GPT-4V. However, we carefully checked all images where the LLMs were unable to locate the pedestrians and observed that 52.94\% (more than half of the sample) are captured during dusk or sunset or within a shady environment or else when there is a large solar glare captured in the image.

% metrics used for each image:
% overlap
% recall
% union
% intersection over union

% \begin{figure}
%     \centering
%     \includegraphics[width=\linewidth]{figures/histogramforaveragerecallacrossrunsperimage.png}
%     \caption{Average recall (across runs) per image}
%     \label{fig:recall}
% \end{figure}

% Blue is 4V, red is 4o
% That in average, it does better.It does not mean that it is always better.

\subsection{Experiment 3: Comparative Evaluation of Prompts} \label{sec:resultsexp3}

%% ALTERNATIVE PROMPT MIGHT MAKE A DIFFERENCE (HYPOTHESIS)
%We hypothesised that the prompt used can affect the performance of GPT in the tasks described in Sec.~\ref{sec:resultsexp1} and~\ref{sec:resultsexp2}. 

% TEST FOR EXP1
In our final experiment, we studied whether changing the prompt could increase the performance for the two tasks above. We refined the prompt based on the metrics reported in Tab.~\ref{tab:exp1confusion} and tested it with GPT-4V. The new prompt emphasised the need to avoid false positives by adding, ``It is very important for you to say `no' if there is no pedestrian close by,'' following the recommendations by Cheng et al.~\cite{emotionprompt}. Experiment 1 was re-run and the recall dropped to 98.18\% since one of the images containing a pedestrian was mislabelled as false negative. This is an undesired result in the intended application context (i.e., missing out on a real pedestrian where we must not). However, the accuracy raised to 88.41\%, given that fewer images were misclassified in total.

% GENERATED ALTERNATIVE PROMPTS
To determine whether a different choice of prompt would also improve the generation of coordinates, we refined also the original prompt  (marked as P1 in Tab.~\ref{tab:prompts}) used for Experiment 2. The goal was to increase the average recall across images and GPT-4V was asked to generate alternative prompts to this end. Two alternative prompts as listed in Tab.~\ref{tab:prompts} were also tested three times for each image in the selected dataset.

\begin{figure}
    \centering
    \includegraphics[width=\linewidth]{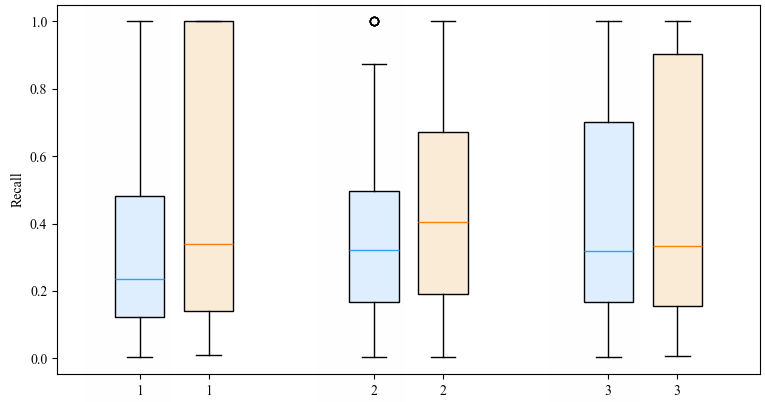}
    \caption{Distribution of recall values among images (for three runs) tested with GPT-4V (blue) and GPT-4o (blue), for the three alternative prompts in Tab.~\ref{tab:prompts}.}
    \label{fig:promptcomparison}
\end{figure}

% THE RESULTS WERE THE SAME, ON AVERAGE
Comparable to prompt P1, the bounding boxes generated by prompt P3 overlapped with the manual ones in 33.82\% of the tests in GPT-4V, whereas it was 43.9\%  for GPT-4o. However, the recall was 47.04\% $\pm$ 34.32\% and 47.6\% $\pm$ 36.21\% on average for GPT-4V and GPT-4o, respectively. These results are a bit higher than that of prompt P1 as seen in Fig.~\ref{fig:promptcomparison}. Prompt P2 also achieved higher average recall values (47.65\% $\pm$ 34.64\% for GPT-4V and 45.01\% $\pm$ 32.35\% for GPT-4o) across all tests. 
Even though there were slight differences in the number of unions going as high as 47.76\% and 45.01\% for P2, the difference is not statistically significant as seen in Fig.~\ref{fig:promptcomparison}, and the high standard deviation highlighted the different performance across images.

\section{Analysis and Discussion} \label{sec:discussion} 

% PROVIDING BACKGROUND INFORMATION: REFERENCE TO THE LITERATURE AND TO THE QUESTION
% motivation / hypothesis: an LLM can act as the hardware abstraction layer in detecting pedestrians
The first question in this study sought to determine how well an LLM can act as a communication interface to detect pedestrians for ADAS. To answer that question, further information about the robustness of LLMs in detecting and locating pedestrians was needed. Therefore, the experiments in this study set out with the aim of assessing whether GPT can detect a pedestrian in an image, as described in Sec.~\ref{sec:resultsexp1}, and describe the area of the image where they are. Further tests as described in Sec.~\ref{sec:resultsexp2} attempted to force the GPT model to use a standard format (e.g., 2D coordinates) to describe the area of an image containing a pedestrian.

% RESTATE RESULTS AND POINT OUT FINDINGS
% RQ1-2: can gpt identify pedestrians? yes! very well (quantitative)
The results of the first experiment determined that GPT-4V is indeed capable of labelling an image whether it contains a pedestrian or not, reaching a recall of 95.24\% and an accuracy of 95.77\%. GPT-4o showcased slightly better results with a near-perfect recall of 99.21\% and an accuracy percentage of 95.45\%. These results show how well the state-of-the-art LLMs perform complex tasks such as pedestrian detection. Within this study, we consider the recall value to be the most crucial metric that indicates the sensitivity of the LLMs. In pedestrian detection, higher recall values indicate fewer false negatives (missed pedestrians), which ultimately helps the system to increase safety by minimizing accidents. 

% RQ2: changing the prompt does not change the results much (but it could if properly done, future work)
Moreover, the accuracy can be further improved by changing the prompt at the expense of a different balance between precision and recall. This might be interesting for scenarios other than the looking-around-the-corner problem, where recall is of the utmost importance.

% RQ2: can gpt tell us where pedestrians are? yes (in natural language)
% RQ2: how well can gpt tell us where pedestrians are? not great (quantitative) because of output format (coordinates)
Another finding that stands out from our results is that while GPT-4V and GPT-4o can provide natural language descriptions about the position and heading of the pedestrian in the images at impressive details, it did not perform well in providing 2D coordinates to describe the bounding boxes around them. In two out of three cases, bounding boxes generated by GPT-4V did not overlap with the area occupied by the pedestrians at all, as reported in Sec.~\ref{sec:resultsexp2}. GPT-4o results were slightly better than GPT-4V where, in an average of 45\% of the tests, the generated bounding boxes overlapped with the area occupied by the pedestrians according to the ground truth data. 

% RQ2: different images and runs get different results
Moreover, our results show that (i) the variability of the recall across images is large, and that (ii) the accuracy is not consistent across different runs using the same image either. 
This undesired behaviour makes GPT-4V and GPT-4o not ready yet to determine the location of pedestrians; however, as reported in the OpenAI API documentation~\cite{gpt4v}, future updates of the model might address this limitation.

% RQ2: changing the prompt does not change the results much (but it could if properly done, future work)
Following this experiment, as presented in Sec.~\ref{sec:resultsexp3}, the prompt was updated and the performance of GPT-4V and GPT-4o was evaluated. The behaviour of each model was arbitrary in each prompt under different runs, and the performances were better on average. However, the null hypothesis could not be rejected, and studying alternative formulations for the prompt is out of the scope of this study.

%\subsection{Threats to Validity}
%\label{sec:threatsToValidity}

We discuss threats to validity based on Feldt and Magazinius \cite{RobertFeldt_ThreatsToValidity}. During the study, we used the LLMs Copilot, DALL.E 2, GPT-4V and GPT-4o, in which we have no control over model-level updates. Had there been any model updates during the experiment period, the results and performance evaluations would have been changed. We consider this as the main internal threat of the study. The selection of the dataset, curating a subset of images for the experiments, and formulating the curation criteria could be considered subjective to the authors' knowledge and might cause a potential bias. Although a preliminary study was conducted with a sample of grey-scale images, it is not evident whether there is a significant difference in accuracy between colour and grey-scale images. Therefore, further studies are required, and results may have an impact based on the image type. In addition to that, technological advancements in the field of automotive, including different sensor technologies and hardware equipment, might limit the generalisability of the results. The ethical concerns, rules, and regulations related to LLMs and Artificial Intelligence may also have an impact on the application level, limiting the generalisability. Therefore we consider them as the external threats that may cause an impact on the study.

\section{Conclusion and Future Work}
\label{sec:conclusion}

% Restating the aims of the study
We have suggested and systematically evaluated a novel approach to enrich a vehicle's understanding of its surroundings to overcome limitations when being dependent on V2I with RSUs or streaming a lot of data in V2V setups. Our proposed concept addresses situations where the ego vehicle's FOV is partially occluded, such as in looking-around-the-corner situations.

Our approach adopts LLM as a communication interface for connected, cooperative (automated) mobility (C-CAM) where the vehicles can have a ``dialogue'' about a traffic situation to abstract from the underlying sensor hardware and IP-sensitive software implementations. Furthermore, practical issues from interoperability would be substantially reduced with an LLM-based abstraction. Our experiments show that LLMs exhibit impressive capabilities to understand fine-granular details of a situation reliably. However, the current generation of LLMs has insufficient performance in locating objects reliably. Nevertheless, we consider the balance of precision and recall as reported in Tab.~\ref{tab:exp1confusion} acceptable in the context of our research goal for this study to reliably detect pedestrians when there are indeed pedestrians present. Other problems, however, might rely on different metrics (e.g., specificity, accuracy, or F1-score) to determine whether GPT-4V and GPT-4o are performing superiorly compared to dedicated ML models.

% Summarising main research findings
%However, the performances should be improved significantly to be able to apply them in real-world use cases. For instance, we would need to work on figuring out where the car providing images is located and where it is facing. 
%As we discuss in Sec.~\ref{sec:discussion}, this balance of precision and recall as reported in Tab.~\ref{tab:exp1confusion} can be regarded acceptable in the light of the research goal of this study to reliably detect pedestrians when there are indeed pedestrians present. Other problems, however, might rely on different metrics (e.g., specificity, accuracy, or F1-score) to determine whether GPT-4V is performing superiorly compared to dedicated ML models.

%Therefore edge computing without sharing heavy and private stuff is possible! So cool for our problem about looking around-the-corner scenes.

% Recognising the limitations of the current study, acknowledging limitation(s) whilst stating a finding or contribution
While LLM-based object detection is in its infancy and has not yet been optimised to answer with exact locations of objects~\cite{gpt4v}, our results show the potential to expect for image detection and localisation. However, further improvements and research work need to be conducted on various complex scenarios that are not addressed yet by this single study, for instance, scenarios where multiple vehicles are prompting, complex road scenarios where multiple pedestrians and cyclists are involved, etc. 
%Hence, the experiment should be extended in a manner that covers versatile scenarios where multiple pedestrians are involved and understand the role of occlusions and carried objects.
% Making recommendations for further research work, setting out recommendations for practice or policy
Summing up to the questions that remain such as to what extent the LLM's performance can scale with the input data and the complexity of the environment. Furthermore, exploring the performance of locally running LLMs needs is left open as well. 

In addition to that, the authors are already working on evaluating the trustworthiness of the proposed pedestrian detection mechanism as a further study focusing on hallucination detection and mitigation. 

%? Why some images are well-interpreted and some not? What is the role of prompt engineering? Role of 3rd party LLMs in edge computing and remote sensing or scene interpretation.
%The reasons why there is such a difference between images is outside the scope of this work and left for future work. 

% Suggesting implications for the field of knowledge, explaining the significance of the findings or contribution of the study
%With the level of advancements in the field of LLMs, we can expect such a stage in the near future. 

% NOTE IMPLICATIONS AND FIVE SUGGESTIONS
% future work: techniques to go from natural language to coordinates using regex and a lot of love.
%CBe
% The proposed methodology below (cf.~Sec.~\ref{sec:experimentDesign}) can be extended as a subsequent study using colour images to evaluate the influence of image type

\balance

\section*{Acknowledgments}
This work has been supported by the Swedish Foundation for Strategic Research (SSF), Grant Number FUS21-0004 SAICOM and the Wallenberg AI, Autonomous Systems and Software Program (WASP) funded by the Knut and Alice Wallenberg Foundation.

\bibliography{reference}
\bibliographystyle{IEEEtran}
\vspace{12pt}

\end{document}